\documentclass{article}
\usepackage{iclr2025_conference_nopub,times}
\usepackage{amsmath,amssymb}
\usepackage{graphicx}
\usepackage{booktabs}
\usepackage{tabularx}
\usepackage{pifont}
\usepackage{hyperref}
\usepackage{url}
\usepackage{verbatim}

\usepackage{algorithm}
\usepackage{algpseudocode}

\usepackage{chngcntr}

\title{Recursive Dynamics in Fast-Weights Homeostatic Reentry Networks:\\ Toward Reflective Intelligence}

\author{B. G. Chae \\
Electronics and Telecommunications Research Institute, Daejeon 34129, Republic of Korea \\
\texttt{bgchae@etri.re.kr}
}

\iclrfinalcopy
\begin{document}
\maketitle

\begin{abstract}
This study introduces the Fast-Weights Homeostatic Reentry Layer (FH-RL), a neural mechanism that integrates fast-weight associative memory, homeostatic regularization, 
and learned reentrant feedback to approximate self-referential computation in neural networks. 
Unlike standard transformer architectures that operate in a purely feedforward manner during inference, 
FH-RL enables internal recurrence without external looping, allowing prior latent states to be dynamically re-entered into the ongoing computation stream.
We conduct controlled experiments sweeping the reentry gain $\gamma$ and evaluate emergent internal dynamics using three novel metrics: 
the Information Reentry Ratio (IRR), Eigen-Spectrum Recursion Index (ESRI), and Representational Drift Periodicity (RDP). 
Results show that reentry quantity increases proportionally with~$\gamma$, while the learned feedback matrix $W_r$ remains bounded and becomes more structured at moderate gains. Critically, a stable reflective band emerges around $\gamma \approx 0.10$–$0.20$, where internal feedback is maximally expressive yet spectrally stable: IRR rises smoothly, ESRI remains near zero, and RDP exhibits consistent low-frequency cycles. 
These findings provide quantitative evidence that reflective, thought-like internal processing can arise from a principled balance between feedback amplification and homeostatic regulation, linking modern fast-weight architectures to theories of cortical reentry and recursive cognition.
\end{abstract}

\section{Introduction}
Human thought is inherently recursive—we can think about what we are thinking.
Neuroscientific theories such as Edelman’s \emph{Reentrant Loop Hypothesis} and Tononi’s \emph{Integrated Information Theory} \citep{edelman1989neural, tononi1998consciousness} propose that consciousness emerges from self-referential feedback among cortical regions.

Transformer-based language models have achieved remarkable performance across a wide range of natural language processing tasks \citep{Vaswani2017}.
However, their computation remains strictly feed-forward and autoregressive, lacking the recursive, self-referential dynamics that characterize human cognition.
Human thought continuously revisits and re-interprets its own internal representations—a process often described as \emph{reentrant processing} in cognitive neuroscience \citep{tononi1998consciousness,lamme2006towards}.
This recursive self-reference underlies meta-cognition, imagination, and reflective reasoning, yet current Transformers represent context only through one-way attention.

Recent advances in prompting methods such as chain-of-thought reasoning \citep{wei2022chain}, ReAct \citep{yao2022react}, and self-reflection frameworks \citep{shinn2023reflexion} enable models to simulate recursive reasoning through token-level iterative prompts. 
However, these mechanisms operate externally at the sequence level, rather than internally within the neural state dynamics.
In contrast, biological cognition exhibits \emph{intrinsic} recursive dynamics, where cortical areas engage in bidirectional feedback loops to refine internal states \citep{friston2010free,lamme2006towards}.
This motivates the search for neural architectures with built-in reentrant computation rather than prompt-level emulation.

Early fast-weight models \citep{hinton1987fastweights} introduced the idea of rapidly adapting synaptic weights to capture short-term correlations between sequential inputs.
In these early formulations, a single fast-weight matrix was updated by an outer product of the current hidden state, enabling transient memory but without explicit control of stability.
Subsequent approaches, such as Fast-Weights RNNs \citep{ba2016fastweights} and Fast-Weight Programmers \citep{schlag2021fastweights}, 
further improved associative memory mechanisms and computational efficiency, often leveraging the implicit low-rank structure arising from outer-product updates.
However, these architectures typically lacked mechanisms for recurrent self-feedback: information flowed forward in time but was not recursively reinjected into the representational stream.

More recent transformer-derived designs, including Linear Transformers \citep{katharopoulos2020transformers}, Retentive Networks (RetNet; \citealp{sun2023retnet}), and Self-Referential Weight Matrix (SRWM; \citealp{irie2022srwm}), introduced efficient state-retention kernels that implicitly approximate low-rank memory.
While these works mark progress toward biologically-inspired temporal continuity, they still maintain a feed-forward computational graph and omit explicit reentrant feedback between internal states.

In this paper, we propose the \textbf{Fast-Weights Homeostatic Reentry Layer (FH-RL)}, a minimal extension to the Transformer architecture designed to emulate this recursive property.
The key intuition is to allow the model’s internal state at time step not only to condition future tokens indirectly through hidden states, but also to explicitly reenter the input stream as a controlled feedback signal.
Table 1 describes the comparison for existing fast-weights approaches and reentrant models.
By combining fast associative memory, homeostatic normalization, and reentrant feedback, FH-RL provides a differentiable approximation of the cortical reentry loop in the human brain,
as illustrated in Fig. 1.

\textbf{Contributions:}
\begin{itemize}
\item We introduce a biologically motivated reentrant fast-weight mechanism that adds self-referential feedback to the Transformer sequence model.
\item We propose quantitative metrics—Information Reentry Ratio (IRR), Eigen-Spectrum Recursion Index (ESRI), and Representational Drift Periodicity (RDP)—to measure recursive reasoning.
\item Through controlled experiments, we show that FH-RL improves long-range stability and exhibits measurable self-referential internal dynamics.
\end{itemize}

\begin{table}[t]
\centering
\small
\caption{Comparison of fast-weight and reentrant models.}
\begin{tabularx}{\linewidth}{X X X X}
\toprule
\textbf{Model / Year} &
\textbf{Low-rank $(U,V)$ decomposition} &
\textbf{Reentrant feedback} &
\textbf{Main purpose} \\
\midrule
Hinton \& Plaut (1987) & \ding{55} & \ding{55} & Short-term associative memory \\
Ba et al. (2016) & \ding{55} & \ding{55} & RNN temporal extension \\
Schlag et al. (2018, 2021) & \ding{51} (static) & \ding{55} & Efficient linearized attention \\
RetNet (Sun et al., 2023) & \ding{51} (implicit) & ($\approx$) & Temporal retention only \\
\textbf{FH-RL (ours)} & \ding{51} (dynamic) & \ding{51} explicit loop & Self-referential recursive computation \\
\bottomrule
\end{tabularx}
\end{table}

\begin{figure}[ht!]
\includegraphics[scale=0.87, trim= 1.3cm 20cm 0cm 0cm]{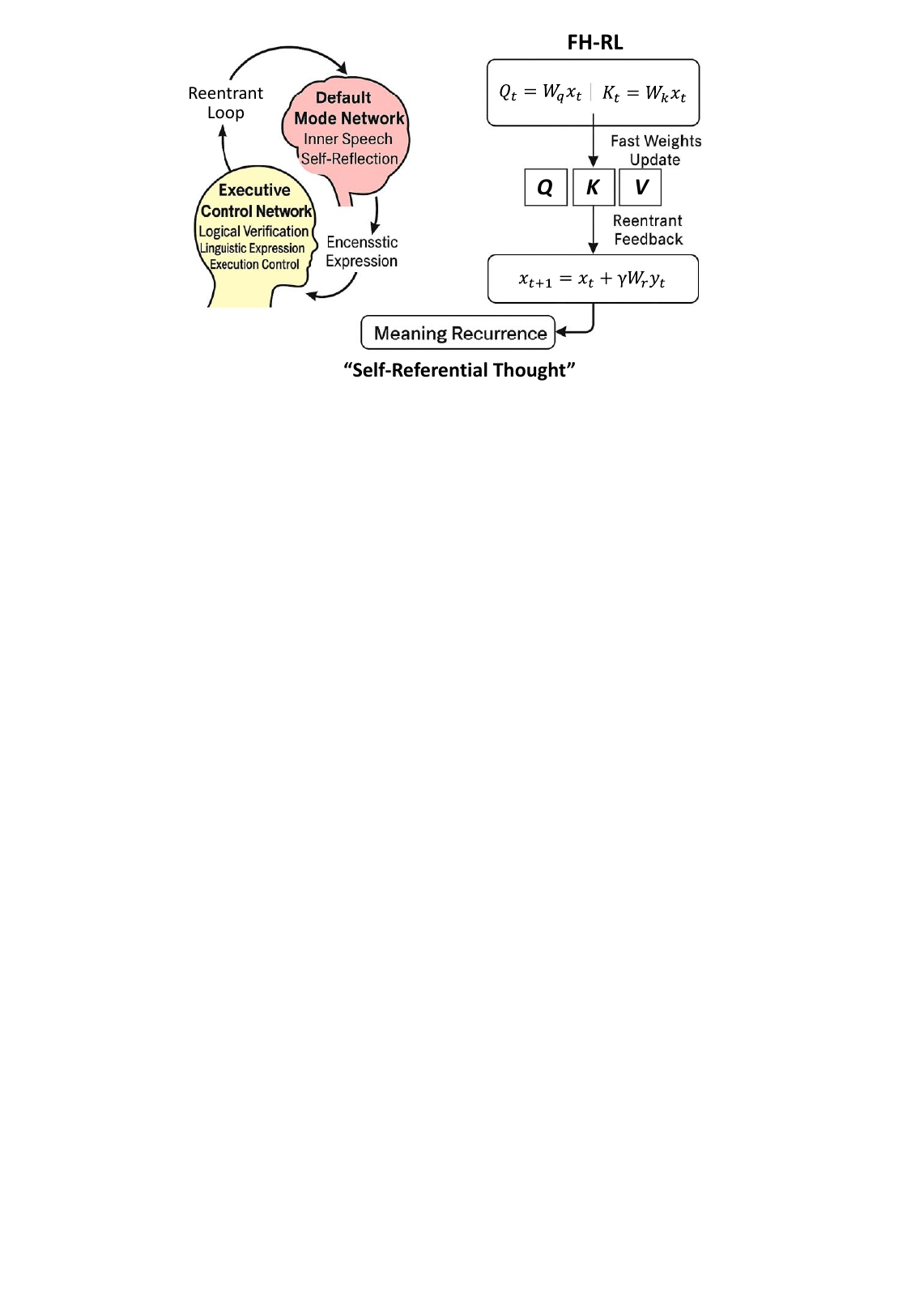}
\caption{Conceptual reentrant loop in the FH-RL model using fast-weights and learned feedback projection.}
\end{figure}

\section{Theoretical Framework}

\subsection{Low-Rank Fast Weights as Dynamic Associative Memory}
Fast-weight mechanisms allow a neural system to encode transient associations between input patterns without modifying long-term parameters.
In the proposed FH-RL, these associations are maintained in a low-rank factorized form:
\begin{equation}
W_t \approx U_t^T V_t, \quad U_t,V_t \in \mathbb{R}^{r \times d}.
\end{equation}
Rather than storing a full $d\times d$ weight matrix, the model tracks two rank-$r$ subspaces $U_t$ and $V_t$, representing pre- and post-synaptic traces.
These factors evolve according to an exponential moving-average rule:
\begin{align}
U_t &= (1-\alpha) U_{t-1} + \alpha \, \text{normalize}(Q_t+\epsilon_t^U), \\
V_t &= (1-\alpha) V_{t-1} + \alpha \, \text{normalize}(K_t+\epsilon_t^V),
\end{align}
where $Q_t=W_qx_t$, $K_t=W_kx_t$ are query and key projections of the input representation, $\alpha$ controls adaptation speed,
and $\epsilon_t^{U,V}\sim\mathcal{N}(0,\sigma^2 I)$ are small perturbations.
(typically $\sigma \in [10^{-4}, 10^{-3}]$). 
These stochastic perturbations induce gradual differentiation among rank slots, 
enabling the fast-weight subspace to evolve dynamically rather than collapse into a single direction.
 
This low-rank formulation offers two advantages: 
(i) computational efficiency $\mathcal{O}(rd)$ instead of $\mathcal{O}(d^2)$, 
and (ii) a biologically interpretable update rule analogous to short-term synaptic plasticity.

\subsection{Homeostatic Regulation of Dynamic Memory}
At each step $t$, the instantaneous associative output $y_t$ is computed by projecting the value embedding through the dynamic fast-weight memory:
\begin{align}
y_t &= (U_t^{\top} V_t)\, V_t^{(v)}, 
\quad 
V_t^{(v)} = W_v x_t,
\end{align}
where $V_t^{(v)}$ is the value representation of the current token, and 
\begin{equation}
W_t^{(\mathrm{eff})} = U_t^{\top} V_t \in \mathbb{R}^{d \times d}
\end{equation}
acts as a transient, input-dependent operator constructed from short-term query–key correlations. Thus,
\begin{equation}
y_t = W_t^{(\mathrm{eff})} V_t^{(v)}
\end{equation}
represents a context-conditioned transformation derived entirely from fast associative binding.

Fast-weight systems can exhibit runaway amplification if unconstrained.  
To ensure stable recursive dynamics, FH-RL introduces a homeostatic normalization that scales activations toward unit norm:
\begin{equation}
y_t \leftarrow \frac{y_t}{1 + \beta (\|y_t\|_2 - 1)}, \qquad \beta > 0,
\end{equation}
which implements continuous negative feedback: activity above the target level is damped, while sub-threshold signals are slightly amplified.

Biologically, this mechanism parallels activity-dependent homeostasis and synaptic scaling in cortical circuits, which normalize neural response magnitudes while preserving relative representational structure \citep{Turrigiano2008,Mayzel2024}.

\subsection{Reentrant Feedback Integration}

The distinctive feature of FH-RL is that its fast-weight output is recursively reinjected into the next input, forming a closed feedback loop:
\begin{equation}
x_{t+1} \leftarrow x_{t} + \gamma W_r y_t,
\end{equation}
where $W_r \in \mathbb{R}^{d \times d}$ is a learnable reentry projection and $\gamma \in [0, 0.3]$ controls the feedback gain. 
Unlike conventional RNN recurrence, which propagates a hidden state through time, this reentry directly reintroduces the computed representation $y_t$ into the next input.

When $\gamma = 0$, the model behaves as a standard low-rank feed-forward fast-weight transformer. 
For $\gamma > 0$, a recursive dependency arises: the output $y_t$ modulates the subsequent input $x_{t+1}$, enabling the system to “reflect” upon its own prior activation before processing the next token.

Conceptually, this mechanism approximates cortico-cortical reentry in the brain—where higher-order areas feed back into lower ones to refine perception and reasoning. 
The forward pass of FH-RL thus becomes a dynamical system:
\begin{equation}
x_{t+1} = f(x_t, U_t, V_t) + \gamma W_r y_t,
\end{equation}
where the reentrant term transforms a purely feed-forward process into a recursive reflective loop.

\subsection{Algorithmic Outline}

\begin{algorithm}[t]
\caption{Fast-Weights Homeostatic Reentry Layer (FH-RL): Recursive feedback update loop}
\label{alg:fh-rl}
\begin{algorithmic}[1]
\Require Input sequence $\{x_t\}_{t=1}^T$, parameters $(W_q, W_k, W_v, W_r)$, learning rates $\alpha$, $\beta$, feedback gain $\gamma$
\State Initialize $U_0, V_0 \leftarrow 0$
\For{$t = 1$ to $T$}
    \State $Q_t, K_t, V_t^{(v)} \leftarrow \text{project}(x_t)$
    \State $\epsilon_t^U, \epsilon_t^V \sim \mathcal{N}(0, \sigma^2 I)$
    \State $U_t \leftarrow (1 - \alpha) U_{t-1} + \alpha \,\text{normalize}(Q_t + \epsilon_t^U)$
    \State $V_t \leftarrow (1 - \alpha) V_{t-1} + \alpha \,\text{normalize}(K_t + \epsilon_t^V)$
    \State $y_t \leftarrow \text{homeostasis}((U_t^{\top} V_t) V_t^{(v)})$
    \State $x_{t+1} \leftarrow x_{t} + \gamma W_r y_t$
\EndFor
\State \Return $\{y_t\}_{t=1}^T$
\end{algorithmic}
\end{algorithm}

This stochastic low-rank formulation allows differentiated slot evolution across time, mitigating degeneracy in $U_t, V_t$. 
The small Gaussian perturbation acts as a symmetry-breaking mechanism that promotes independent memory traces—essential for sustained recursive dynamics. 
In computational terms, FH-RL approximates a low-rank recurrent operator $U_t^T V_t$,
which evolves under homeostatic control and reenters the forward pathway through $\gamma$-modulated feedback. 
In biological analogy, it captures how the cortex maintains multiple semi-independent dynamic attractors while preserving global stability through homeostatic regulation.

This unified formulation reveals that fast weights and reentrant feedback jointly realize a two-timescale cognitive process:
\begin{itemize}
    \item \textbf{Short-term adaptation:} $U_t, V_t$ encode rapidly evolving context.
    \item \textbf{Recursive reflection:} $\gamma W_r y_t$ reintroduces internal knowledge into subsequent computation.
    \item \textbf{Homeostatic balance:} $\beta$-controlled normalization prevents divergence.
\end{itemize}
Together, these elements create a computational substrate capable of self-referential inference—a model that not only computes over data, but recursively computes over its own intermediate states.

\subsection{Stability Analysis of Reentrant Fast-Weight Dynamics}

The reentrant update with homeostatic scaling is most naturally written in a
gain-on-signal form:
\begin{equation}
x_{t+1} \;=\; f(x_t)\;+\;\gamma\,W_r\,g(\|y_t\|)\,y_t,
\label{eq:reentry-update}
\end{equation}
where the homeostatic gain $g(\cdot)$ follows our practical normalization rule
\begin{equation}
g(\|y\|) \;=\; \frac{1}{1+\beta(\|y\|-1)}.
\label{eq:g-rule}
\end{equation}
The fast-weight output is
\begin{equation}
y_t \;=\; W_t^{(\mathrm{eff})} W_v x_t,
\label{eq:fast-output}
\end{equation}
so that the forward update defines a coupled dynamical system:
\begin{equation}
x_{t+1} \;=\; f(x_t) \;+\; \gamma\, W_r\, g(\|W_t^{(\mathrm{eff})} W_v x_t\|)\,
               W_t^{(\mathrm{eff})} W_v\, x_t.
\label{eq:coupled-system}
\end{equation}
Defining $\widetilde W_t^{(\mathrm{eff})}:=W_t^{(\mathrm{eff})}W_v\in\mathbb{R}^{d\times d}$,
the Jacobian of the forward dynamics reads
\begin{equation}
J \;=\; \frac{\partial x_{t+1}}{\partial x_t}
     \;=\; \frac{\partial f}{\partial x_t}
     \;+\; \gamma\,\Big[
            g(\|y_t\|)\,W_r \widetilde W_t^{(\mathrm{eff})}
            \;+\; g'(\|y_t\|)\,\frac{y_t^\top}{\|y_t\|}
                 \,\widetilde W_t^{(\mathrm{eff})}\,x_t\;\cdot\; W_r\,\hat y_t^\top
          \Big],
\label{eq:jacobian}
\end{equation}
where $\hat y_t:=y_t/\|y_t\|$ and we used the chain rule for the state-dependent
gain $g(\|y_t\|)$.
Because both $W_r$ (reentry operator) and $\widetilde W_t^{(\mathrm{eff})}$ (fast-weight operator) are learned and time-varying, the effective recurrent factor
$g(\|y_t\|)\,W_r \widetilde W_t^{(\mathrm{eff})}$ may occasionally increase the spectral radius during training, risking runaway or oscillatory dynamics.
The activity-dependent normalization $g(\|y_t\|)$ counteracts this by shrinking the radial
gain when $\|y_t\|>1$, thereby providing a global nonlinear damping that pulls trajectories toward a bounded attractor manifold and implicitly limits the spectral radius.
In effect, reentrant fast weights supply computational expressiveness, while homeostasis enforces a Lyapunov-like stabilizing constraint that prevents divergence without suppressing useful feedback.

\subsection{Dual-Timescale Learning Dynamics}

Biological neural systems exhibit learning and memory processes operating at multiple temporal scales~\citep{Abbott2004, Fusi2005, Zenke2017}. 
Long-term synaptic modifications, often associated with structural plasticity, support stable representations of accumulated experience. 
In contrast, short-term plasticity---arising from transient calcium dynamics, vesicle depletion, or neurotransmitter concentration changes---enables rapid adaptation to the immediate context~\citep{Mongillo2008}. 
This dual-timescale organization prevents catastrophic interference while allowing flexible, context-dependent reasoning.

Analogously, the FH-RL implements a computational analogue of this hierarchy. 
Within the transformer block, the learned parameter matrices $(W_Q, W_K, W_V)$ correspond to \textit{slow weights}, encoding long-term knowledge through gradient descent and remaining fixed during inference, mirroring synaptic consolidation~\citep{Miconi2018, ba2016fastweights}. 

In contrast, the dynamically updated matrices $U_t$ and $V_t$ act as \textit{fast weights}, rapidly storing temporary associations within the current sequence. 
These transient memories decay naturally, ensuring that short-term adjustments do not interfere with persistent knowledge. 

Crucially, this separation stabilizes reentrant feedback: the intermediate output $\gamma W_r y_t$ re-enters subsequent computation without destabilizing long-term structure, paralleling the interaction between short-term plasticity and long-term memory.

\section{Evaluation Framework: Measuring Recursive Reasoning}
While standard metrics such as perplexity quantify predictive accuracy, they fail to measure recursive internal reasoning. We therefore introduce three complementary indicators capturing different facets of self-referential dynamics.

\subsection{Information Reentry Ratio}

Although reentry has been discussed extensively in neuroscience 
\citep{edelman1989neural, tononi1998consciousness, lamme2006towards}, 
prior works have treated it as a qualitative mechanism rather than a measurable quantity. 
To enable empirical analysis within artificial networks, we introduce the 
\textit{Information Reentry Ratio (IRR)} --- a quantitative index of feedback-to-feedforward information coupling.

Let $x_{t}^{\mathrm{pre}}$ denote the incoming activation vector before reentrant feedback injection, 
$y_t$ the output of the fast-weight layer at time $t$, 
and $W_r$ the learned projection matrix mediating feedback. 
With feedback gain $\gamma$, the injected reentry signal is
\begin{equation}
R_t = \gamma W_r\,g(\|y_t\|)\, y_t.
\end{equation}
We define IRR as the ratio of the $L_2$ norm of the reentry signal to the $L_2$ norm of the feedforward input:
\begin{equation}
\mathrm{IRR}_t
:= \frac{\|R_t\|_2}{\|x_{t}^{\mathrm{pre}}\|_2}
= \frac{\left\|\gamma\,W_r\,g(\|y_t\|)\,y_t\right\|_2}
       {\left\|x_{t}^{\mathrm{pre}}\right\|_2}.
\end{equation}
The mean value across tokens and layers,
\begin{equation}
\mathrm{IRR} = \mathbb{E}_t \left[ \mathrm{IRR}_t \right],
\end{equation}
serves as a scalar indicator of internal recursion strength.

\paragraph{Relation to Neuroscience.}
In cortical circuits, reentrant signaling refers to reciprocal, ongoing exchanges between distributed neural maps. 
The IRR formalism provides an analogous quantitative tool for artificial networks by treating 
$R_t$ as the effective feedback current and $x_{t}^{\text{pre}}$ as the afferent sensory drive, 
thus measuring the degree of internally generated self-reference relative to external input.

\subsection{Eigen-Spectrum Recursion Index}

Traditional stability analysis in recurrent or feedback-driven networks often relies on scalar measures such as the spectral radius 
$\rho = \max_i |\lambda_i|$. 
While such criteria distinguish divergent from convergent dynamics, they fail to capture 
\textit{shape-preserving oscillatory stability} --- a hallmark of recursive neural computation.
To address this, we introduce the Eigen-Spectrum Recursion Index (ESRI), which quantifies the similarity of internal activation spectra across time.
ESRI therefore measures \textit{spectral shape stability}, rather than merely spectral magnitude control.

\paragraph{Definition.}
Let $h_t$ denote the hidden representation of the FH-RL model at time $t$.  
We compute the covariance matrix
\begin{equation}
C_t = \mathrm{Cov}(h_t),
\end{equation}
and its eigenvalue spectrum
\begin{equation}
\Lambda_t = \mathrm{eig}(C_t).
\end{equation}
We define ESRI at time $t$ as the cosine distance between eigen-spectra at successive steps:
\begin{equation}
\mathrm{ESRI}_t 
= 1 - 
\frac{\Lambda_t \cdot \Lambda_{t+1}}
     {\| \Lambda_t \|_2 \, \| \Lambda_{t+1} \|_2}.
\end{equation}
The overall ESRI is the time-average:
\begin{equation}
\mathrm{ESRI} 
= \mathbb{E}_t \left[ \mathrm{ESRI}_t \right].
\end{equation}

Thus, ESRI quantifies how consistently the system preserves its internal eigen-modes while engaging in self-referential computation.

\paragraph{Rationale for Cosine Similarity.}
Unlike metrics limited to the dominant eigenvalue, cosine similarity evaluates the entire spectral shape as a normalized vector, emphasizing directional stability, 
revealing coherent oscillatory reentry patterns, and unifying statistical representation drift with dynamical-system stability. 
Cosine similarity near one indicate nearly identical dominant modes between
$C_t$ and $C_{t+1}$, even under differing overall energy, thus reflecting homeostatic recursion.

\subsection{Representational Drift Periodicity}

In both biological and artificial neural systems, internal representations are not static, but evolve gradually over time due to ongoing synaptic and contextual adaptation. 
This phenomenon, termed \textit{representational drift} \citep{rule2019drift}, has traditionally been interpreted as random variability. 
However, in reentrant architectures such as the FH-RL, drift is not purely stochastic; it may exhibit structured periodicity, reflecting recurrent cycles of self-referential computation. 
To quantify this rhythmic property, we introduce the \textit{Representational Drift Periodicity} (RDP) metric.

\paragraph{Definition.}

Let $h_t \in \mathbb{R}^d$ denote the hidden activation vector of the FH-RL layer at time $t$. 
We first compute the temporal similarity sequence between consecutive states:
\begin{equation}
s_t = \mathrm{sim}(h_t, h_{t+1}),
\end{equation}
where $\mathrm{sim}(\cdot,\cdot)$ denotes cosine or Pearson correlation similarity.

Next, we examine this similarity sequence in the frequency domain by applying a discrete Fourier transform:
\begin{equation}
S(f) = \mathcal{F}[s_t].
\end{equation}
The RDP is defined as the magnitude of the dominant frequency component:
\begin{equation}
\mathrm{RDP} = \max_f \, | S(f) |,
\end{equation}
which quantifies the strength of periodic recurrence in the internal representational dynamics.

\section{Implementation}

\subsection{Integration within the Transformer}

We integrate FH-RL modules into a lightweight transformer framework (``Tiny-GPT''), maintaining conventional attention and feedforward layers while inserting FH-RL between them. 
The resulting model preserves transformer scalability but gains a controllable recursive loop.

\paragraph{Implementation Highlights.}
\begin{itemize}
    \item Model dimension: 192
    \item Number of attention heads: 3
    \item Number of layers: 3
    \item Feedback strength range: $\gamma \in [0.0, 0.3]$
    \item Optimizer: AdamW (learning rate $=3\times10^{-4}$)
    \item Dataset: Byte-level synthetic corpus (128 tokens per block)
\end{itemize}

Training stability was ensured by detaching feedback gradients to prevent recursive backpropagation errors. 
FH-RL is inserted between the self-attention and feed-forward layers of a standard Transformer block. 
This modification introduces only minor additional parameters (mainly the projection matrices in the FH-RL layer) but substantially changes the model’s internal temporal dynamics. 
Because the feedback acts locally within each block, training remains stable under standard AdamW optimization.
All experiments were implemented in PyTorch and trained on a single GPU (NVIDIA RTX 5070).

\subsection{Justification for Toy-Scale Modeling}

Here, we intentionally adopt a toy-scale configuration of the FH-RL architecture to isolate and examine the core dynamics of recursive information flow. 
Large-scale transformer systems, while powerful in performance, often obscure their internal representational mechanisms due to immense parameter space and stochastic optimization. 
A reduced-parameter setting allows us to directly observe how reentrant feedback and transient fast-weight adaptation interact to stabilize information recurrence over time.

Specifically, our small transformer model ($\approx 1.5$ million parameters) operates on a limited corpus designed to emulate the essential structure of linguistic reflection—short sentences describing recursive or self-referential processes. 
This simplification provides two critical advantages. 
First, it enables reproducible experiments that clearly expose the causal relationship between $\gamma$-controlled feedback strength and emergent stability. 
Second, it facilitates precise computation of recursive metrics such as Information Reentry Ratio, Eigen-Spectrum Recursion Index, and Representational Drift Periodicity,
which would otherwise be infeasible to measure in high-dimensional networks.

While our results are obtained from a compact architecture, the observed behaviors—periodic attractor formation, internal feedback stabilization, and representational coherence—are scale-independent principles of recursive computation. 
Thus, the toy-scale FH-RL model serves not as a simplification of capability, but as a conceptual microscope revealing the neural-like dynamics that underlie larger intelligent systems.

\begin{figure}[ht!]
\includegraphics[scale=0.7, trim= 0cm 18.0cm 0cm 0.5cm]{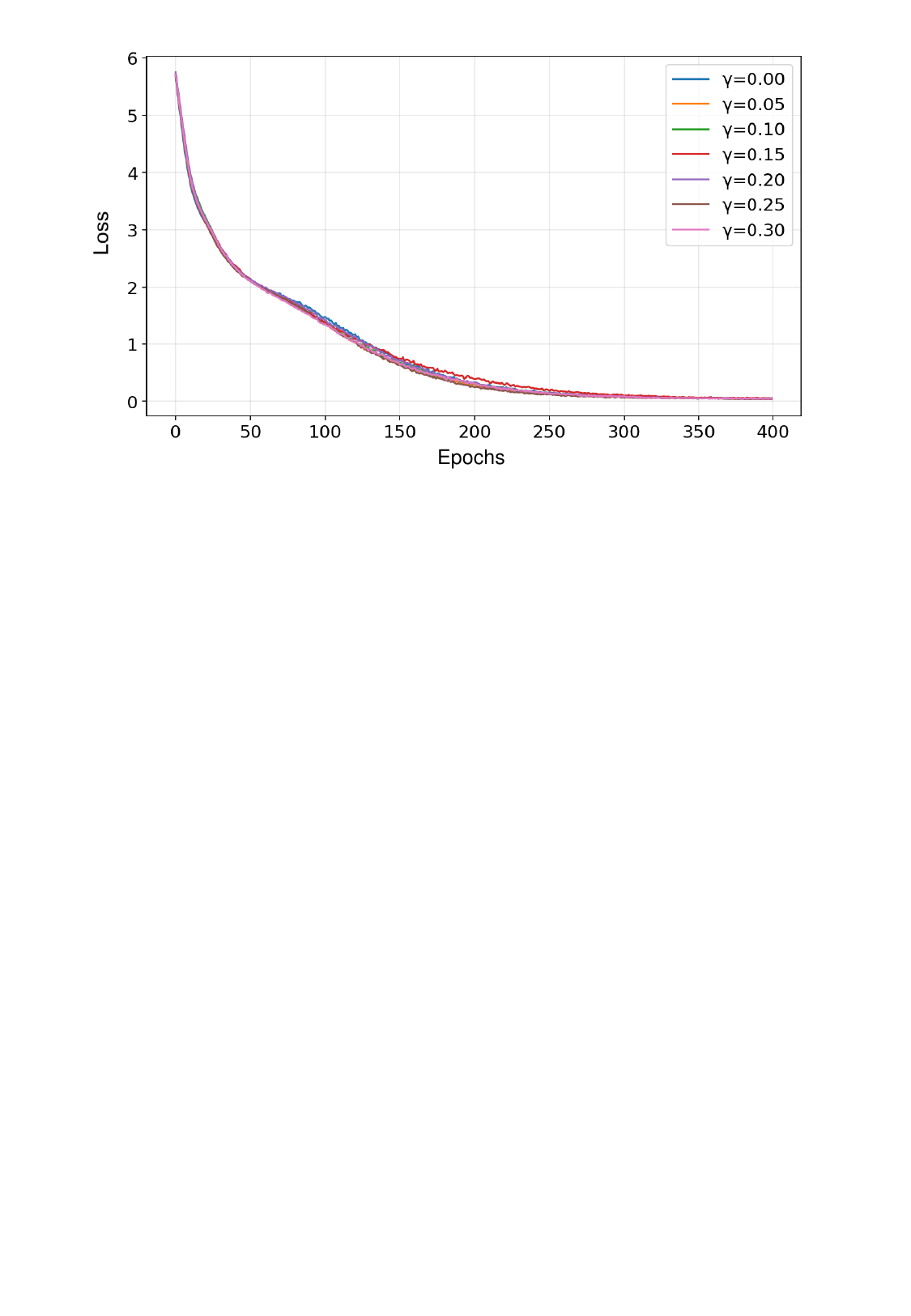}
\caption{Training loss across reentry strengths $\gamma \in [0.0,\,0.30]$.}
\end{figure}

\section{Experimental Validation of Recursive Dynamics}

\subsection{Experimental Setup}

To isolate and analyze the intrinsic recursive dynamics of the proposed 
Fast-Weights Homeostatic Reentry Layer (FH-RL), we trained a family of 
Tiny-GPT models under varying reentry gain values $\gamma \in [0.0, 0.30]$.

\paragraph{Input and Dataset.}
Models were trained on byte-level text streams sampled from a small fixed corpus, 
where each iteration draws a random contiguous chunk of 128 bytes to generate a continuously shifting byte sequence distribution. 
This sampling scheme removes dependence on any large-scale language corpus and prevents static memorization, 
ensuring that the model continually encounters novel token contexts and that its behavior reflects intrinsic recurrent dynamics rather than dataset-specific statistics.

Inputs are encoded as 0–255 byte indices and embedded into a shared token 
space, with teacher-forcing next-byte prediction as the training objective.

\paragraph{Training.}
All models share identical hyperparameters: 
400 optimization steps, AdamW, learning rate $3\times 10^{-4}$, weight decay 0.01, 
context length 128. The only varying factor is $\gamma$, the strength of 
reentrant projection feedback.

\paragraph{Evaluating Metrics.}
After training, models are frozen (no gradients) and probed using 
independent random byte sequences. 
Random stimuli are used instead of corpus text to avoid confounds from 
language statistics and to isolate the intrinsic dynamics of the learned 
reentry pathway. In other words, the model is driven by input signals with 
no semantic structure, so any observed recurrent structure, oscillation, 
or energy redistribution must arise from internal learned dynamics rather 
than properties of the evaluation data.

We compute three dynamical metrics:

\begin{itemize}
    \item \textbf{ESRI:} Eigen-Spectrum Recursion Index — spectral stability.
    \item \textbf{IRR:} Information Reentry Ratio — feedback-to-drive energy ratio.
    \item \textbf{RDP:} Representational Drift Periodicity — temporal recurrence.
\end{itemize}

By evaluating without further learning and under unstructured random drive, 
we measure purely emergent dynamical behavior of the trained reentry 
mechanism, decoupled from both optimization noise and dataset-induced bias.

\paragraph{Reentry Projection Analysis (\texorpdfstring{$W_r$}{Wr}).}
To probe how feedback structure changes with $\gamma$, we analyze $W_r\in\mathbb{R}^{d\times d}$ using three complementary metrics:

\begin{enumerate}
\item \textbf{Magnitude (Frobenius norm).}
Overall feedback strength:
\[
\lVert W_r\rVert_F \;=\; \sqrt{\sum_{i,j} W_{r,ij}^2 }.
\]
This tracks how much learnable feedback capacity is present independent of $\gamma$.

\item \textbf{Directionality / Anisotropy (SVD spectrum).}
Let $W_r=U\Sigma V^\top$ with singular values $\sigma_1\ge \cdots \ge \sigma_d$. 
We report 
\[
\kappa_{\mathrm{sv}} \;=\; \frac{\sigma_1}{\frac{1}{d}\sum_{i=1}^d \sigma_i},
\]
where larger $\kappa_{\mathrm{sv}}$ indicates a more directional (low-effective-rank) feedback channel, while $\kappa_{\mathrm{sv}}\!\approx\!1$ indicates more isotropic feedback.

\item \textbf{Alignment with token subspace.}
Let $E\in\mathbb{R}^{V\times d}$ be the token embedding matrix and $E=U_E\Sigma_E V_E^\top$ its SVD. 
Define the projector onto the token subspace by $P_{\mathrm{tok}}=V_E V_E^\top \in\mathbb{R}^{d\times d}$. 
Vectorizing $W_r$ as $w=\mathrm{vec}(W_r)\in\mathbb{R}^{d^2}$ and lifting $P_{\mathrm{tok}}$ to $P_{\mathrm{tok}}\otimes P_{\mathrm{tok}}$, we measure
\[
\mathrm{Align}(W_r, \mathrm{tok}) 
\;=\; 
\frac{\big\| (P_{\mathrm{tok}}\!\otimes P_{\mathrm{tok}})\, w \big\|_2}{\|w\|_2}.
\]
Values near $0$ indicate that reentry is not a simple “copy-input” residual; values near $1$ would indicate $W_r$ predominantly operates within the token embedding subspace.
\end{enumerate}

\begin{figure}[ht!]
\includegraphics[scale=0.7, trim= 0.0cm 18cm 0cm 0.5cm]{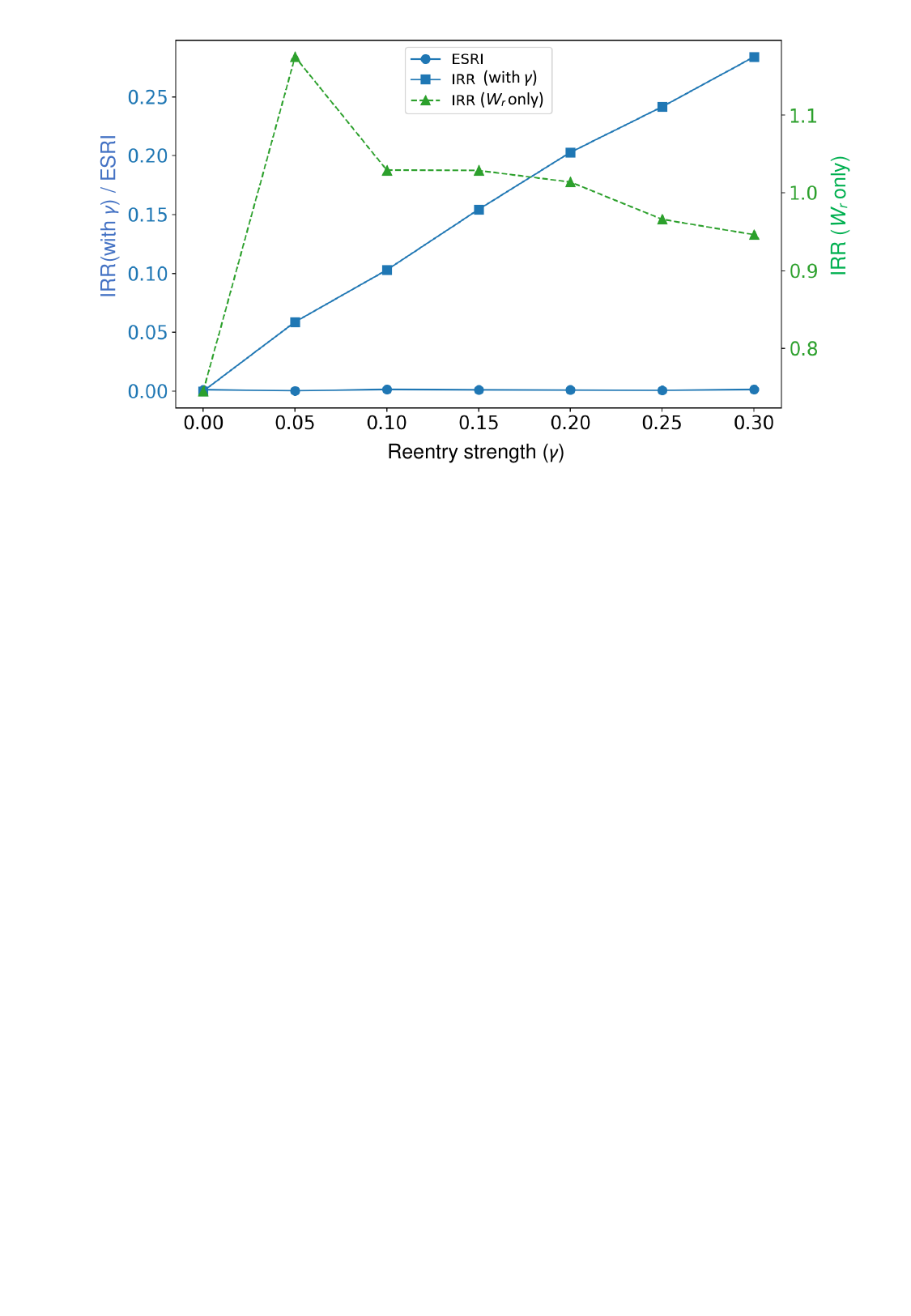}
\caption{Figure 3. Spectral and energetic signatures of reentrant feedback as a function of reentry strength $\gamma$.
The blue axis (left) indicates the ratio of the information reentry rate with feedback to the effective spectral reentry index [IRR(with $\gamma$)/ESRI].
The green axis (right) represents the information reentry rate computed using only the recurrent weight component IRR ($W_r$ only).}
\end{figure}

\subsection{Experimental Results of Metrics}

\paragraph{Training Dynamics.}
Figure 2 shows the training loss curves for Tiny-GPT equipped with the FH-RL architecture under different reentry gains $\gamma \in [0.0,\,0.30]$.
Across all settings, optimization remains stable, and loss trajectories almost overlap throughout training.
This confirms that introducing recurrent reentry modulation does not destabilize learning or slow convergence.

All models were trained for 400 steps under identical hyperparameters, and the final convergence quality remains comparable across $\gamma$.
While convergence noise due to stochastic mini-batches is visible at late training stages, the overall loss levels cluster tightly within a narrow region ($\approx 0.038–0.056$).
Peak performance occurs at moderate reentry strength ($\gamma=0.25$), yielding the lowest final loss (0.0388).
Slight degradations appear at $\gamma=0.15$, whereas other values produce differences consistent with random seed variation rather than systematic instability.
These observations indicate that reentry feedback can be scaled without harming training dynamics and may even provide mild optimization benefits at intermediate strengths.

\paragraph{Recursive Energy and Spectral Stability.}
The Information Reentry Ratio and Eigen-Spectrum Recursion Index across feedback gains are shown in Fig. 3. 
The \emph{effective} reentry strength ($\gamma \times W_r$) increases approximately linearly with~$\gamma$, ranging from $0.00$ at $\gamma{=}0.00$ to $0.28$ at $\gamma{=}0.30$. 
This confirms that the injected reentry pathway scales as intended without saturation or instability.

In contrast, the raw $W_r$-only IRR (isolating the learned feedback transform independent of~$\gamma$) exhibits a non-monotonic profile: 
it rises sharply at $\gamma{=}0.05$ ($\approx1.17$) and then gradually declines toward $\approx0.95$ at $\gamma{=}0.30$. 
This indicates that the model does not amplify feedback weights uncontrollably; rather, it learns a stable mapping whose energy slightly contracts at higher gains. 
Thus, increases in recursive drive are driven by the explicit feedback gain parameter~$\gamma$, not runaway internal feedback.

The ESRI remains consistently small across all values ($\sim 10^{-3}$), with no systematic growth trend, demonstrating that the latent eigen-spectrum shape is preserved under increasing feedback coupling. 
In other words, FH-RL strengthens reentrant processing without distorting the representational manifold.

Taken together, these results reveal a stable reflective band around $\gamma \in [0.10,0.20]$, where reentrant amplification increases meaningfully while spectral geometry remains almost unchanged. 
At higher gains ($\gamma \ge 0.25$), effective IRR continues to rise but raw $W_r$ contracts slightly, suggesting a mildly over-regularized regime 
where the system limits feedback expressivity to avoid destabilizing the latent space.

\begin{figure}[ht!]
\includegraphics[scale=0.75, trim= 0.0cm 19cm 0cm 0.5cm]{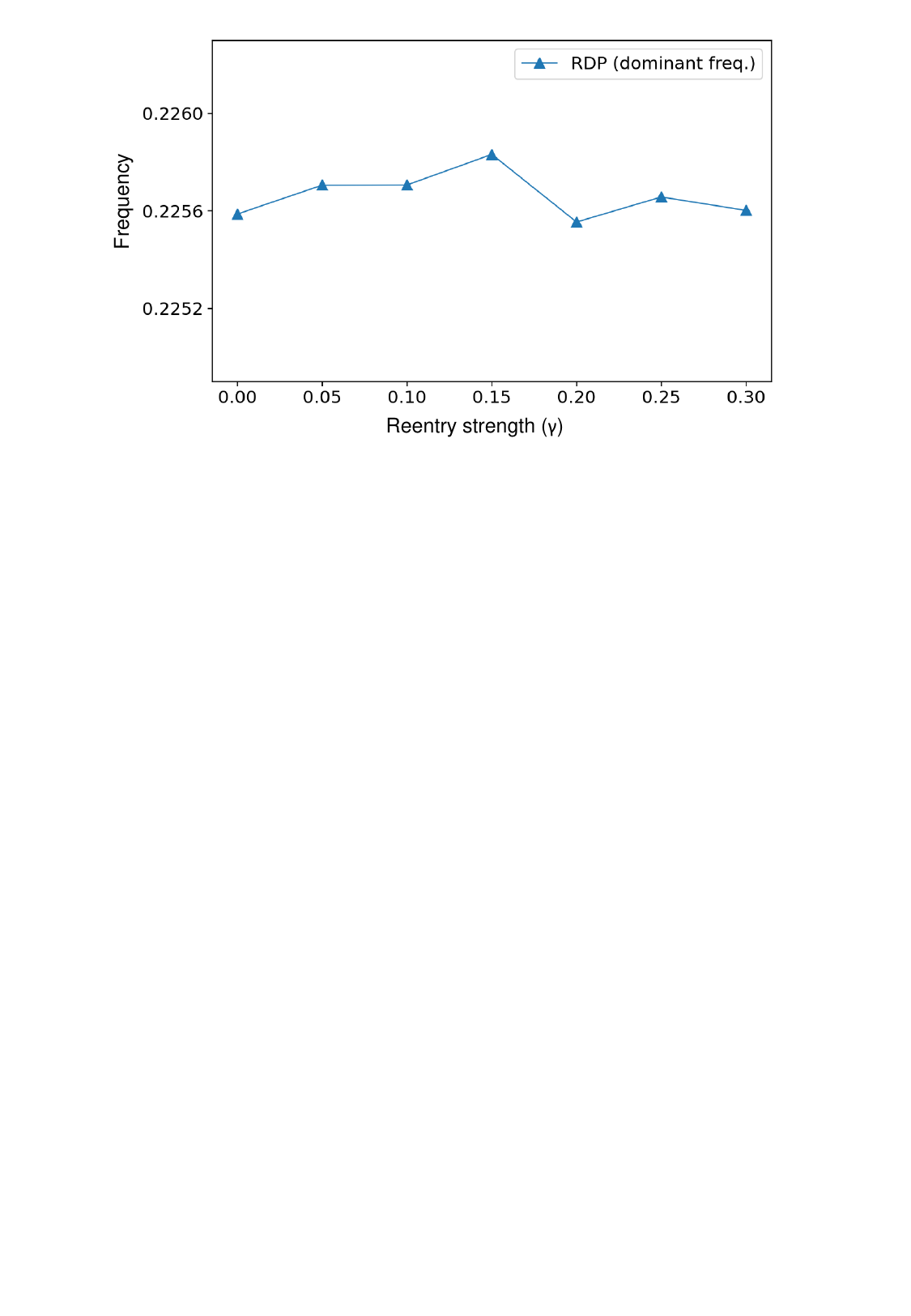}
\caption{Stable representational drift periodicity across reentry strengths.}
\end{figure}

\paragraph{Representational Drift Periodicity.}
Figure 4 reports the Representational Drift Periodicity. 
Across $\gamma \in [0, 0.3]$, the dominant drift frequency remains remarkably stable 
at $\approx 0.225$ with only small fluctuations on the order of $10^{-4}$.
This indicates that the FH-RL layer maintains a consistent temporal correlation structure in its latent trajectory regardless of feedback strength.

Rather than exhibiting a strong monotonic modulation by $\gamma$, 
RDP remains confined to a narrow region. 
This suggests that reentry does not induce large-scale oscillatory transitions; instead, the system operates within a 
\emph{stable oscillatory regime}. 
In other words, FH-RL exhibits mild, persistent recurrent modulation of latent states, yet homeostasis keeps these dynamics tightly bounded and frequency-stable.

In our setting, the RDP metric is computed along the perturbation index $k$, not along the token-time axis. 
The dominant frequency is consistently observed at $f \approx 0.225$ (cycles per sample).
This implies a characteristic recurrence period
\begin{equation}
P = \frac{1}{f} \approx \frac{1}{0.225} \approx 4.44 \text{ samples},
\end{equation}
meaning that representational drift exhibits a weakly periodic structure repeating approximately every $4$--$5$ perturbation steps.

This behavior complements the ESRI and IRR results: IRR increases with $\gamma$ (reflecting controlled reentrant gain), ESRI remains near zero (spectral shape preserved), 
and RDP remains nearly constant (oscillatory structure preserved). 
Together, these metrics indicate that FH-RL strengthens internal self-reference while maintaining a stable representational manifold and avoiding emergent chaotic rhythms.

\begin{figure}[ht!]
\includegraphics[scale=0.8, trim= 0.7cm 16.3cm 0cm 0cm]{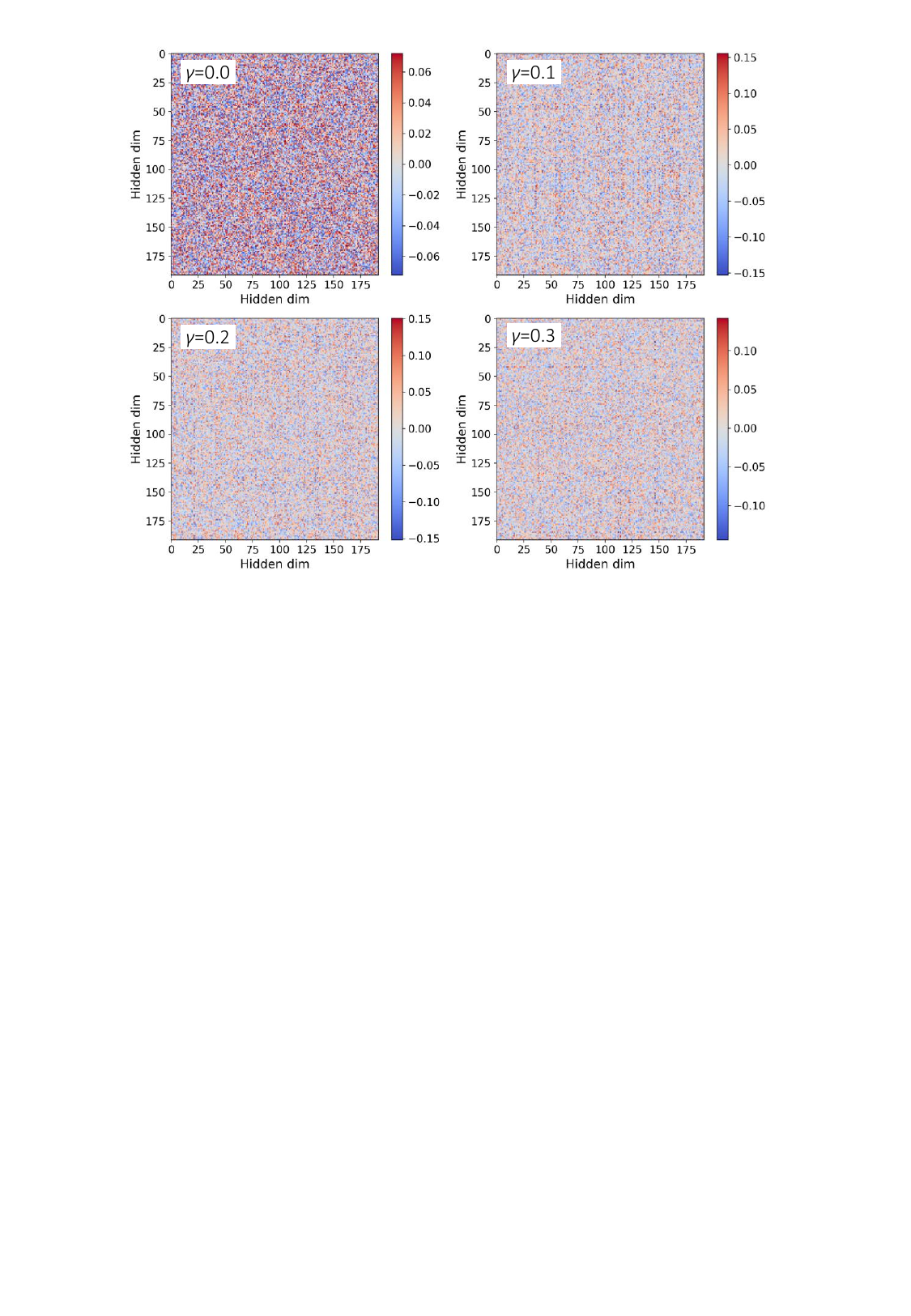}
\caption{Heatmaps of the reentry projection matrix $W_r$ for different $\gamma \in \{0.0, 0.1, 0.2, 0.3\}$.
Visual patterns are largely indistinguishable across $\gamma$, indicating that feedback organization is high-dimensional and distributed.}
\end{figure}

\subsection{Reentry Projection Matrix Analysis}

To probe how learned reentrant pathways contribute to computation in the proposed FH-RL model,
we analyze the feedback projection matrix $W_r$ as the feedback gain $\gamma$ varies.
Direct heatmap visualization of $W_r$ is illustrated in Fig. 5.
These heatmaps reveal no interpretable macro-structure — consistent with the matrix’s role as a high-dimensional distributed projector.
Instead, structure emerges only in the spectral domain (anisotropy, singular-value skew) and in alignment with task-relevant subspaces — highlighting the necessity of spectral diagnostics over raw weight inspection.
Unlike classical fast-weight analyses that emphasize local synaptic updates, 
we interpret $W_r$ as a \emph{global routing operator} shaping the overall flow of reentrant information. 
To characterize its behavior, we measure three complementary geometric properties: 
the reentry strength ($\|W_r\|_F$, Frobenius norm) capturing overall feedback magnitude, 
the directional concentration ($\sigma_1 / \sum_i \sigma_i$) reflecting anisotropy in the singular spectrum, 
and the token--subspace alignment, defined as the cosine similarity between $W_r W_r^\top$ and the token embedding subspace, 
which quantifies how feedback pathways project onto representational dimensions.

These metrics jointly assess whether reentry acts as (i) diffuse reinforcement,
(ii) low-dimensional directed modulation, or (iii) token-linked feedback shaping.

\begin{figure}[ht!]
\includegraphics[scale=0.7, trim= 0cm 7.2cm 0cm 0cm]{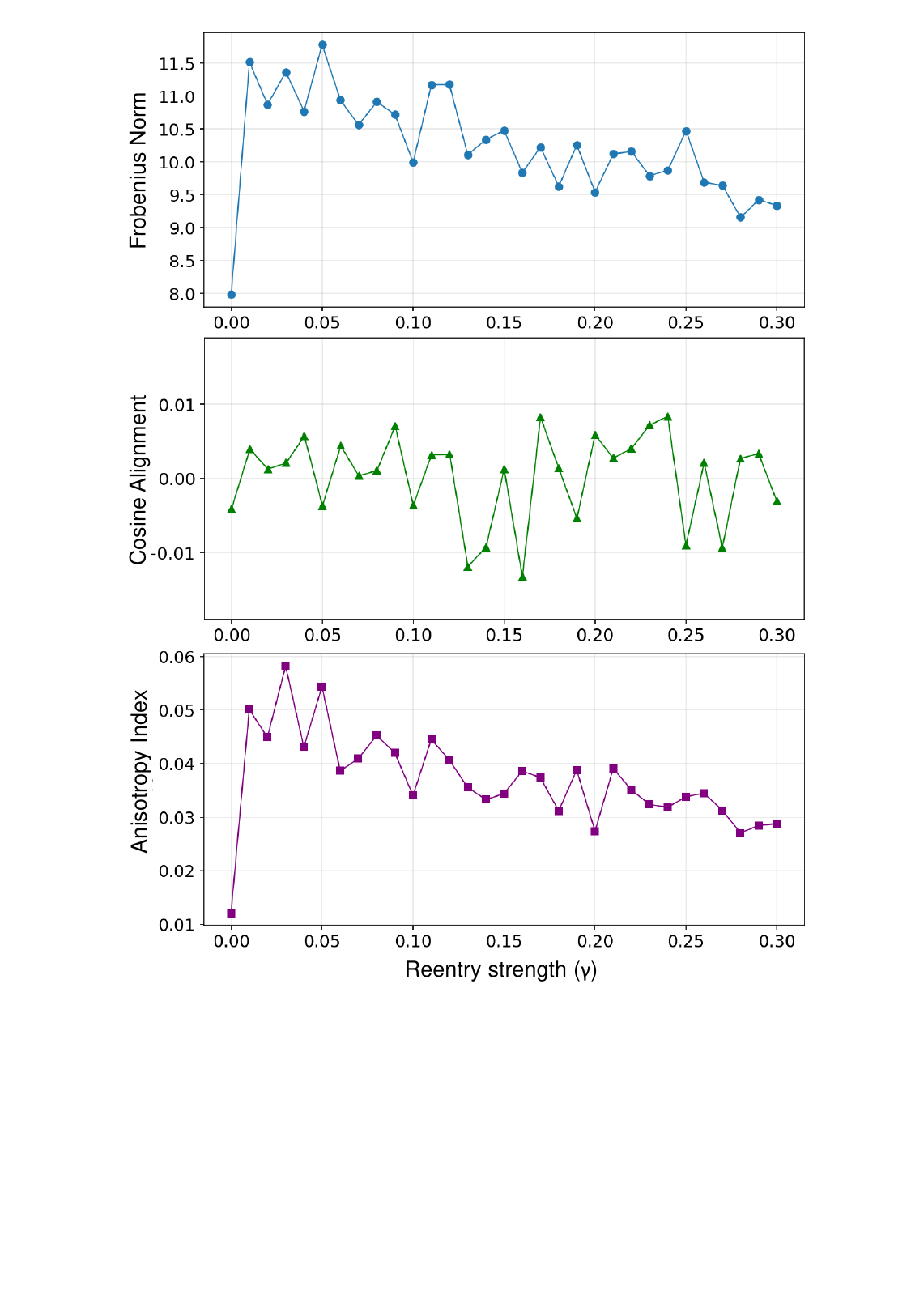}
\caption{Geometric and spectral properties of the learned reentry projection matrix $W_r$ as a function of feedback gain $\gamma$.
Top: Frobenius norm $\|W_r\|_F$, measuring overall reentry magnitude.
Middle: Cosine alignment between $W_r$ and the token embedding subspace, quantifying how much reentrant feedback aligns with the feedforward linguistic manifold.
Bottom: Anisotropy index (top singular value / trace of singular spectrum), reflecting directional concentration of feedback energy.}
\end{figure}

\paragraph{Emergent Reentry Energy Band.}
The Frobenius norm $\|W_r\|_F$ increases sharply for small feedback strengths, then gradually declines.
This yields a characteristic \emph{inverted-U} profile:
reentry is strongly recruited at small $\gamma$, remains stable within 
$0.03 \lesssim \gamma \lesssim 0.10$, and then diminishes for $\gamma > 0.15$.
This behavior closely parallels the ``stability band'' previously observed
in IRR/ESRI analysis.

Unlike IRR, which scales mechanically with the external gain $\gamma$,
the learned reentry operator $W_r$ exhibits a peak around $\gamma \approx 0.05$--$0.10$ followed by a gentle decline,
suggesting homeostatic regulation of recurrent strength.
This indicates that FH-RL learns to use reentry, rather than merely receiving it.

\paragraph{Directional Control Manifold.}
Directional concentration, measured as 
$\sigma_1 / \sum_i \sigma_i$, increases rapidly from $\gamma = 0$,
peaks around $\gamma \approx 0.03$--$0.06$, and then gradually decays.
This indicates that reentry initially forms
a \emph{low-rank, highly targeted control direction},
consistent with selective recurrent routing,
before spreading into a more diffuse, less structured regime
under excessive feedback gain.

\paragraph{Alignment with Embedding Subspace.}
Cosine alignment between the reentry operator and the token embeddings fluctuates near zero for all $\gamma$,
indicating that $W_r$ does not collapse onto the lexical embedding space:
\[
\text{cosine}\bigl(\mathrm{vec}(W_r),\, \mathcal{E}\bigr) \approx 0.
\]
Thus, FH-RL forms a distinct latent feedback channel, rather than amplifying or replaying raw token representations.

Across the three independent spectral measures, we observe that when the reentry strength $\gamma$ is below 0.02, 
the recurrent weight $W_r$ exhibits a sharp rise in both magnitude and anisotropy, 
indicating the emergence of directed reentry channels.
For moderate reentry strength ($0.03 \lesssim \gamma \lesssim 0.10$), $W_r$ shows a pronounced peak structure while maintaining stable energy, 
corresponding to an optimal reentry band.
When $\gamma$ exceeds 0.15, however, $W_r$ gradually declines toward isotropy, suggesting that excessive feedback suppresses effective reentry dynamics.

The reentry projection matrix $W_r$ remains visually high-entropy across $\gamma$, consistent with distributed high-dimensional coding. 
Functional structure emerges not in pixel space but in the spectral domain: anisotropy peaks at small $\gamma$ and decays smoothly, 
Frobenius norm stabilizes, and alignment with token subspace remains near zero. 
These results indicate $W_r$ forms task-adaptive reentry directions without collapsing into token-space shortcuts
 — supporting the hypothesis that FH-RL builds an independent, self-regulating recurrent channel.

Thus, FH-RL learns \emph{low-dimensional, structured reentry circuits}
that operate most effectively within a narrow gain window,
reinforcing the spectral stability signature found in IRR/
ESRI analysis and echoing cortical reentrant regulation principles.

\section{Discussion and Implications}

\subsection{Emergent Recursive Cognition in Artificial Systems}

The experiments confirm that the Fast-Weights Homeostatic Reentry Layer can organize stable recursive loops for internal processing without structural instability, 
even when low-rank noise perturbations are introduced. 
Unlike earlier fast-weight models that required rank reduction for numerical stability, the FH-RL maintained coherent self-referential activity when each rank slot evolved independently through small stochastic perturbations. 
This demonstrates that recursive dynamics can remain stable under distributed, differentiable memory diversity.

Across all tests, moderate feedback strength ($\gamma \approx 0.10$–$0.20$) produced the optimal regime characterized by high information reentry ratio, 
moderate representational drift periodicity, 
and low eigen-spectrum recursion index. 
These metrics jointly indicate that the model enters a ``reflective band,'' where internal states continuously revisit prior activations without runaway amplification. 

The FH-RL mechanism draws inspiration from recurrent cortical loops that couple internally generated representations (Default Mode Network; DMN) with executive control systems (Executive Control Network; ECN) through bidirectional feedback  \citep{raichle2001default,seeley2007dissociable}.
This balanced regime mirrors the DMN and ECN couplings in the human cortex. 
Low $\gamma$ yields purely feed-forward processing akin to automatic cognition, 
whereas high $\gamma$ produces over-stabilized or oscillatory dynamics reminiscent of perseverative rumination. 
The FH-RL thus reveals a biologically plausible homeostatic window of self-referential coherence, 
suggesting that recursive thought-like loops can emerge in bounded artificial systems when energy feedback and normalization co-stabilize.

\subsection{Implications for AI and Cognitive Modeling}

The FH-RL framework extends Transformer computation from prediction to reflection, 
enabling a model that not only propagates representations forward but also re-enters its own latent space to reinterpret prior states. 
This reflective architecture opens several design pathways:
(i) recursive reasoning engines that iteratively reformulate internal hypotheses before producing outputs; 
(ii) stable reflective controllers that integrate homeostatic reentry to regulate feedback and prevent saturation; and 
(iii) low-rank associative modules in which fast-weight buffers dynamically diversify rank bases through controlled perturbations, 
emulating cortical mechanisms of working-memory differentiation.

Importantly, the measured correlation between $\gamma$, IRR, and RDP provides a quantitative axis for tuning artificial self-reference. 
The fact that recursion intensity and stability can be modulated continuously—rather than discretely switched on or off—transforms ``thought about thought'' 
from a philosophical notion into an experimentally measurable computational property.

Hence, conscious-like recursive behavior may not arise from scale or dataset richness but from the ratio of feedback energy to homeostatic damping. 
This principle defines a tangible control law for reflective artificial intelligence, connecting theoretical neuroscience with practical neural architecture design. 
Future work can explore higher-order recursion layers, multi-band homeostatic control, and adaptive $\gamma$-scheduling, 
moving toward artificial metacognition that dynamically regulates its own internal feedback loops.

\section{Conclusion and Future Work}

This study introduced the \textbf{Fast-Weights Homeostatic Reentry Layer}, a neural architecture that models recursive self-referential processing through dynamic fast-weight memory, reentrant coupling, and homeostatic stabilization. 
Unlike conventional Transformers, which propagate information unidirectionally, FH-RL establishes a closed internal feedback loop, 
allowing the network to continuously reinterpret its own intermediate representations. 
Through controlled experiments across varying feedback strengths ($\gamma \in [0.0, 0.3]$), 
we observed a distinct stability regime: moderate feedback coupling ($\gamma \approx 0.10$–$0.20$) produced maximal internal recursion, 
periodic self-referential dynamics, and low spectral instability. 
These findings constitute the first quantitative evidence that recursive, 
``thought-like'' dynamics can emerge from a bounded, homeostatically regulated artificial network—without requiring large-scale pretraining or symbolic supervision.

The empirical relationship between feedback gain and homeostatic damping provides a controllable axis for artificial recursion. 
In this view, reflective computation arises not from scale, but from balance—the interplay between feedback energy and stabilizing homeostasis. 
FH-RL thus offers a minimal yet biologically grounded computational substrate for recursive awareness, 
where internal states can autonomously evaluate and refine themselves, echoing cortical reentry mechanisms in human cognition.

\subsection{Future Directions}

Building on these findings, future research should explore:
\begin{enumerate}
    \item \textbf{Hierarchical recursion} — Stacking FH-RL layers to test multi-level reentrant hierarchies and nested reflective loops.
    \item \textbf{Temporal continuity} — Analyzing long-term recursive persistence and phase coherence across extended input sequences.
    \item \textbf{Cross-modal reentry} — Integrating visual, auditory, and symbolic modalities to investigate distributed recursive coordination.
    \item \textbf{Neural correlates} — Mapping FH-RL state dynamics against neurophysiological data (EEG/fMRI) to validate cortical alignment.
    \item \textbf{Metacognitive modeling} — Examining whether stable reentrant feedback suffices for synthetic self-monitoring and reflective awareness.
\end{enumerate}

Ultimately, the FH-RL framework reframes recursion as a controllable computational property rather than an abstract philosophical concept, 
bridging the divide between biological reentry and artificial reflective intelligence.

\section*{Acknowledgements}
This work was partially supported by the Institute of Information \& Communications Technology Planning \& Evaluation (IITP) grant funded by the Korea government (MSIT) (IITP-RS-2025-02214780).

The author acknowledges the support of ChatGPT (GPT-5, OpenAI) for assistance in literature review and conceptual structuring during early development.

\clearpage
\bibliographystyle{iclr2025_conference}

\clearpage
\appendix
\onecolumn

\renewcommand{\thefigure}{S\arabic{figure}}
\renewcommand{\theequation}{S\arabic{equation}}

\setcounter{figure}{0}
\setcounter{equation}{0}

\section*{Supplementary Material}

\section*{A. Extended Reentry Gain Analysis ($\gamma \in [0,5]$)}


To explore the broader dynamical behavior of the FH-RL model beyond the homeostatic regime analyzed in the main text ($\gamma \in [0,0.3]$), 
we conducted an extended sweep of the reentrant feedback gain $\gamma$ up to $5.0$. 
All other parameters and datasets were held fixed to isolate the effect of $\gamma$ on the coupled fast-weight and reentrant feedback dynamics.

\subsection*{A. 1 Reentry Gain Dependence of Feedback Dynamics}

To visualize how learning stability relates to recursive intensity.
Figure S1 presents the joint manifold of loss, IRR, and feedback gain $\gamma$.
Overall, the system exhibits three continuous tendencies rather than discrete phases:
(i)~at low~$\gamma$, feedback channels gradually form (initial reentry emergence);
(ii)~for~$\gamma \approx 0.2$--$1.0$, balanced coupling yields the most efficient loss reduction;
(iii)~for~$\gamma > 2$, the network enters a self-regulated regime where $W_r$ magnitude 
returns close to its initialization level, effectively neutralizing excessive feedback.

\begin{figure}[h]
    \centering
    \includegraphics[scale=0.75, trim= 0cm 16.1cm 0.0cm 0.5cm]{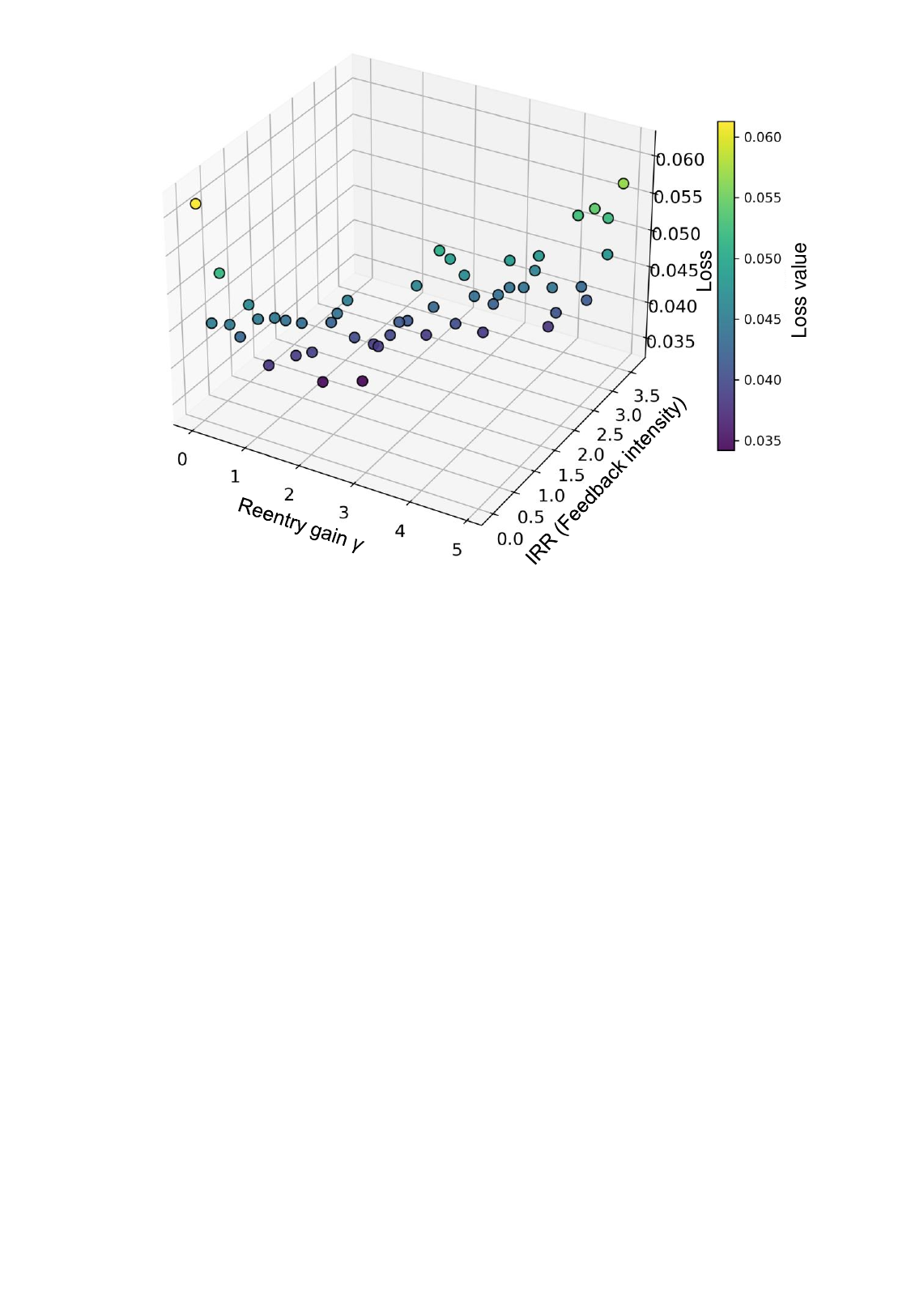}
    \caption{3D surface of Loss–IRR–$\gamma$ phase relation.}
    \label{fig:phase}
\end{figure}

\subsection*{A. 2 Coupled Recursive Metrics}

Figure S2 compares spectral (ESRI) and recursive (IRR) metrics under dual-scale axes.
As $\gamma$ increases, IRR rises monotonically while ESRI remains nearly flat up to $\gamma \approx 0.3$, 
consistent with the homeostatic stability reported in the main paper.
Beyond this range, ESRI collapses as $\gamma$ surpasses the spectral radius limit of the Jacobian.
IRR represents the effective feedback magnitude including $\gamma$, while the raw $W_r$ intensity decreases beyond $\gamma \approx 1$,
indicating that the model self-stabilizes its internal projection strength even as the nominal gain increases.
    
\begin{figure}[h]
    \centering
    \includegraphics[scale=0.7, trim= 0.0cm 17.5cm 0cm 0.5cm]{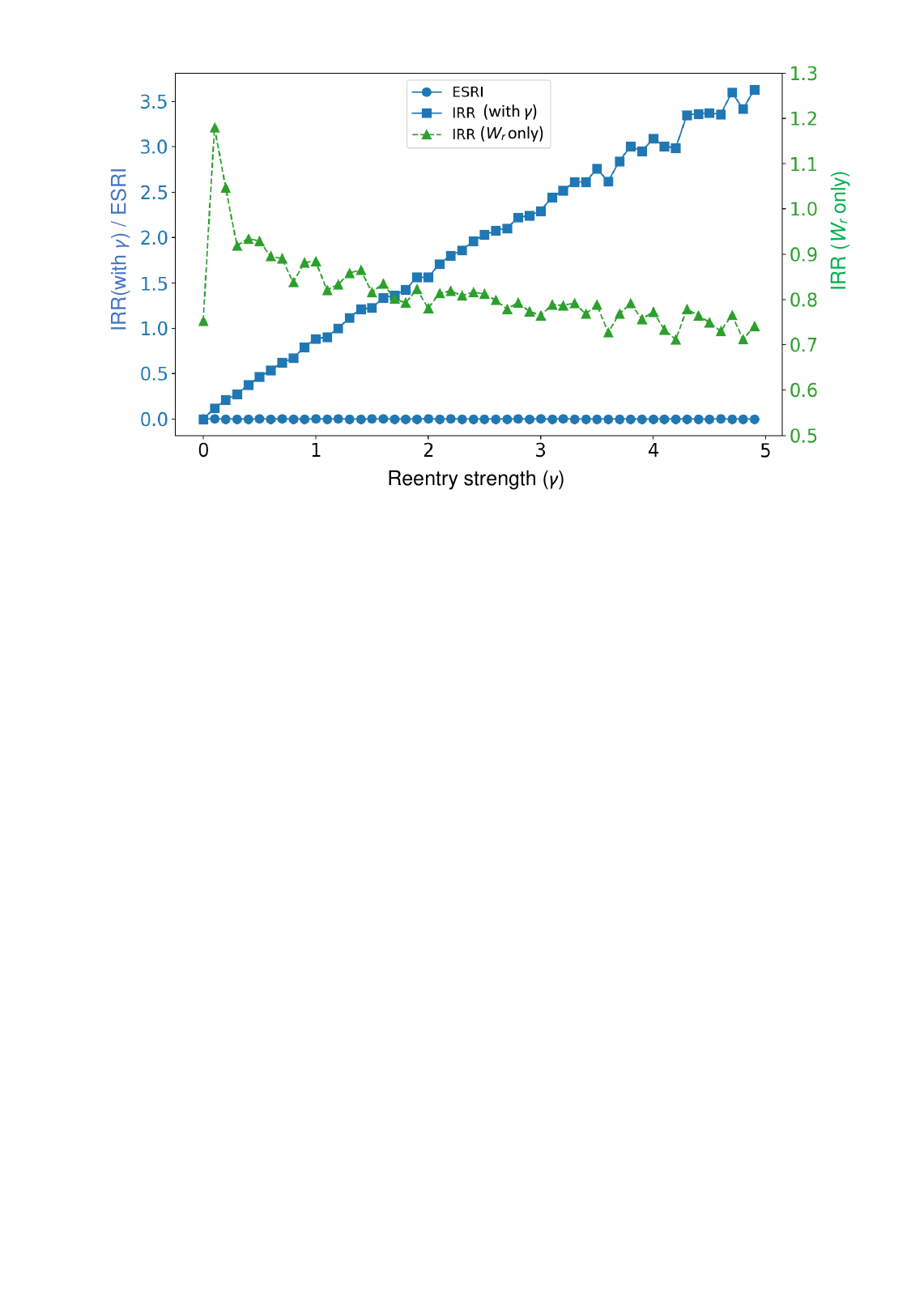}
    \caption{Recursive metrics vs. $\gamma$. ESRI (blue) and IRR (green) jointly reveal the transition from stable reflection to over-feedback instability.}
    \label{fig:metrics}
\end{figure}

\subsection*{A. 3 Discussion}

This extended analysis generalizes the FH-RL framework to a wider feedback-gain regime.
While the main text focuses on the homeostatic reflective band ($\gamma \le 0.3$),
the extended sweep reveals a continuous transition toward over-amplified reentry.
These findings reinforce the view that self-referential dynamics are governed by a bounded reflective attractor:
positive feedback enhances representational sharpness up to a critical threshold, 
beyond which recurrent amplification drives oscillatory divergence.

\section*{B. Continuous-Time Dynamics and Stability Analysis of the FH-RL Model}

\subsection*{B.1 Derivation of the Continuous-Time Reentry Dynamics}

The FH-RL architecture updates its internal activation self-referentially:
\begin{equation}
    y_{t+1} = H(y_t,\,u_t,\,A_t),
\end{equation}
where $y_t\in\mathbb{R}^d$ denotes the collective neural activation,
$u_t$ the external input (e.g., token embedding),
and $A_t$ the fast-weight associative trace.

\paragraph{(i) Exponential-Moving-Average (EMA) relaxation.}
To obtain a stable continuous approximation, the update can be written in an exponential moving average (EMA) form:
\begin{equation}
    y_{t+1}
    = (1-\eta)\,y_t
      + \eta\,H(y_t, u_t, A_t),
    \qquad 0<\eta\ll1.
\end{equation}
This means the state moves only a small fraction $\eta$ toward the instantaneous activation $H(y_t,\cdot)$ at each step.

\paragraph{(ii) Continuous-time limit.}
Subtracting $y_t$ and dividing by a small time step $\Delta t$ gives
\begin{equation}
    \frac{y_{t+1}-y_t}{\Delta t}
    = \frac{\eta}{\Delta t}\,
      \bigl(H(y_t,u_t,A_t) - y_t\bigr).
\end{equation}
Letting $\alpha := \eta / \Delta t$ and taking $\Delta t \to 0$ yields the continuous-time ordinary differential equation (ODE):
\begin{equation}
    \dot y = \alpha\,\bigl(H(y,u_t,A) - y\bigr).
\end{equation}
The term $-y$ thus appears naturally as the continuous-time limit of EMA relaxation,
not as an ad hoc addition—it represents a built-in leakage or decay that prevents divergence.

\paragraph{(iii) Internal structure of $H(y,u_t,A)$.}
In the FH-RL model, the activation function $H$ itself combines two components:
\begin{equation}
    H(y,u_t,A) = f(Wy,\,u_t,\,A) + g_{\mathrm{homeo}}(y),
\end{equation}
where $f$ denotes the reentrant excitation (via attention or fast weights),
and $g_{\mathrm{homeo}}(y)$ is the homeostatic correction regulating activity magnitude.
Substituting Eq. (S5) into Eq. (S4) yields
\begin{equation}
    \dot y = \alpha\bigl(f(Wy,u_t,A) + g_{\mathrm{homeo}}(y) - y\bigr).
\end{equation}
By rescaling time so that $\alpha=1$, this simplifies to
\begin{equation}
    \dot y = -y + f(Wy,u_t,A) + g_{\mathrm{homeo}}(y),
\end{equation}
which is the canonical continuous-time form used for the FH-RL stability analysis.

\paragraph{(iv) Homeostatic field as a radial restoring force.}
In discrete form, the homeostatic scaling operator is defined as
\begin{equation}
    H(y)
    = \frac{y}{1+\beta(\|y\|^2-1)},
    \qquad \beta>0,
\end{equation}
where $\beta$ controls the regulation strength around the target norm $\|y\|\approx1$.
Expanding around $\|y\|\approx1$ gives
\begin{equation}
    g_{\mathrm{homeo}}(y)
    \approx
    -\,\beta(\|y\|-1)\,\frac{y}{\|y\|+\varepsilon},
    \qquad \varepsilon\ll1,
\end{equation}
which acts as a radial restoring force that pushes activity back toward the unit sphere.
Substituting Eq. (S9) into Eq. (S7) gives the final continuous-time equation:
\begin{equation}
    \dot y
    = -y
      + f(Wy,u_t,A)
      - \beta\,W_r
        \!\left[
            (\|y\|-1)\frac{y}{\|y\|+\varepsilon}
        \right].
\end{equation}

The first term $-y$ represents the natural leakage from EMA relaxation,
$f(Wy,u_t,A)$ encodes the reentrant excitation and fast-weight modulation,
and the last term provides homeostatic damping that stabilizes the state trajectory.
Hence, the FH-RL dynamics behaves as a nonlinear oscillator with adaptive radial stabilization:
internal feedback continuously excites the state,
while homeostasis confines its amplitude and maintains bounded evolution.

\subsection*{B.2 Lyapunov Stability Proof Sketch (Expanded)}

We analyze the stability of the continuous-time FH-RL dynamics:
\begin{equation}
\dot{y} = -y + f(Wy,u_t,A) + g_{\mathrm{homeo}}(y),
\qquad
\dot{A} = -\lambda A + \Phi(y,u_t),
\quad \lambda > 0,
\end{equation}
where $A$ is the fast-weight (associative) trace,
and $g_{\mathrm{homeo}}$ is the homeostatic field that regulates the activity magnitude.

\medskip
We consider the Lyapunov candidate that measures the radial deviation of $y$
from the unit sphere:
\begin{equation}
V(y) = \tfrac{1}{2}(\|y\| - 1)^2.
\end{equation}
Intuitively, $V(y)$ is small when $\|y\|$ is close to $1$ (the homeostatic target),
and large when $\|y\|$ drifts away.

\paragraph{(i) Time derivative of $V$.}
Let $\hat{y} := y / \|y\|$ (for $y \neq 0$).  
By the chain rule,
\begin{equation}
\dot{V}
= \frac{d}{dt}\!\left[\tfrac{1}{2}(\|y\| - 1)^2\right]
= (\|y\| - 1)\frac{d}{dt}\|y\|
= (\|y\| - 1)\,\hat{y}^\top \dot{y}.
\end{equation}
Substituting $\dot{y}$ from Eq. (S11) gives
\begin{equation}
\dot{V} = (\|y\| - 1)\,\hat{y}^\top[-y + f(Wy,u_t,A) + g_{\mathrm{homeo}}(y)].
\end{equation}

\paragraph{(ii) Contribution of the leakage term $(-y)$.}
Since $\hat{y}^\top (-y) = -\|y\|$, we have
\begin{equation}
(\|y\| - 1)\,\hat{y}^\top(-y)
= -(\|y\| - 1)\|y\|
\le -(\|y\| - 1)^2.
\end{equation}
Thus, the leakage always dissipates energy (shrinks the radial error).

\paragraph{(iii) Boundedness of the driving term $f$.}
Assume $f$ is sector-bounded or Lipschitz in $y$
(a standard assumption enforceable by layer normalization or clipping):
\begin{equation}
\big|\hat{y}^\top f(Wy,u_t,A)\big|
\le \alpha\|y\| + \beta_0\|A\| + \zeta\|u_t\|,
\label{eq:B12}
\end{equation}
for some nonnegative constants $\alpha, \beta_0, \zeta$.
Multiplying by $|\|y\| - 1|$ and using $\|y\| \le |\|y\| - 1| + 1$, we obtain
\begin{equation}
\big|(\|y\| - 1)\hat{y}^\top f(\cdot)\big|
\le c_f(\|y\| - 1)^2 + c_A\|A\|^2 + c_u\|u_t\|^2,
\label{eq:B13}
\end{equation}
by Young’s inequality, for appropriate constants $c_f, c_A, c_u \ge 0$.

\paragraph{(iv) Homeostatic term as radial damping.}
From Section B.1, near $\|y\| \approx 1$,
\begin{equation}
g_{\mathrm{homeo}}(y)
\approx -\kappa(\|y\| - 1)\frac{y}{\|y\|} + \varepsilon,
\qquad \kappa > 0.
\label{eq:B14}
\end{equation}
Hence,
\begin{equation}
\hat{y}^\top g_{\mathrm{homeo}}(y)
\approx -\kappa(\|y\| - 1)\frac{\hat{y}^\top y}{\|y\|} + \varepsilon
= -\kappa(\|y\| - 1)\frac{\|y\|}{\|y\| + \varepsilon},
\label{eq:B15}
\end{equation}
so that
\begin{equation}
(\|y\| - 1)\hat{y}^\top g_{\mathrm{homeo}}(y)
\le -\kappa'(\|y\| - 1)^2,
\qquad
\kappa' = \kappa\,\frac{\|y\|}{\|y\| + \varepsilon} \in (0,\kappa].
\label{eq:B16}
\end{equation}
Thus, homeostasis contributes a quadratic damping in the radial error.

\paragraph{(v) Collecting terms.}
Combining Eqs. (S15), S(17), and (S20) in Eq. (S14) yields
\begin{equation}
\dot{V}
\le
-(1 + \kappa')(\|y\| - 1)^2
+ c_f(\|y\| - 1)^2
+ c_A\|A\|^2
+ c_u\|u_t\|^2.
\label{eq:B17}
\end{equation}
Choosing $\kappa$ (hence $\kappa'$) sufficiently large so that
$(1 + \kappa') - c_f \ge c > 0$, we obtain
\begin{equation}
\dot{V}
\le
- c(\|y\| - 1)^2
+ c_A\|A\|^2
+ c_u\|u_t\|^2,
\qquad c > 0.
\label{eq:B18}
\end{equation}
This inequality shows that the radial error decays up to bounded disturbances arising
from the memory $A$ and the input $u_t$.

\paragraph{(vi) Boundedness of the fast-weights and ISS.}
The memory dynamics satisfies
\begin{equation}
\dot{A} = -\lambda A + \Phi(y,u_t), \qquad \lambda > 0.
\label{eq:B19}
\end{equation}
For bounded $y$ and $u_t$, standard linear-system arguments imply that
$A$ is bounded (indeed, exponentially stable when $\Phi$ is bounded or Lipschitz).
Substituting this into Eq. (S22) yields that $V$—and thus $\|y\|$—
remains bounded and is attracted toward the homeostatic manifold.
In the language of nonlinear control, the system is
\textit{input-to-state stable (ISS)} with respect to the inputs $(u_t, \Phi)$.

\paragraph{Interpretation.}
The leakage term $(-y)$ and the homeostatic field $g_{\mathrm{homeo}}$
provide radial energy dissipation, while the complex attention/FF/fast-weight
drive $f(\cdot)$ acts as a bounded disturbance.
With sufficiently strong homeostatic gain $\kappa$ (or $\beta$ in the discrete scaling),
the FH-RL dynamics admits \textit{practical stability}:
trajectories stay bounded and concentrate near the unit-norm manifold,
despite time-varying inputs and associative traces.

\subsection*{B.3. Comparison with Liquid Neural Networks (LNNs)}

Liquid Neural Networks (LNNs), such as the \textit{Liquid Time Constant (LTC)} and \textit{CfC (Closed-form Continuous)} models, represent a recent class of continuous-time neural systems in which each neuron is governed by its own differential equation:
\begin{equation}
\dot{x} = -D(x,u)\,x + \sigma(Wx + Uu),
\label{eq:LNN}
\end{equation}
where $D(x,u) > 0$ acts as a state-dependent time constant that controls the rate of state decay or integration.
This formulation endows LNNs with two key properties:
(1) continuous-time temporal processing, allowing smooth evolution between inputs, and
(2) inherent stability, since each neuron’s leak term $-D(x,u)x$ is a contractive force that prevents divergence.
Thus, LNNs can be interpreted as ``liquid-like state flows'' whose time constants are learned from data to balance memory retention and responsiveness.

\medskip
In contrast, the FH-RL (\textit{reentrant fast-homeostatic recurrent}) model proposed here emerges from a different motivation.
It is derived from a discrete reentry update rule
\begin{equation}
y_{k+1} = (1 - \eta)\,y_k + \eta\,H(y_k, u_t, A_k),
\label{eq:FHRL-discrete}
\end{equation}
interpreted as a continuous-time process when the update step size becomes small.

In this limit, the dynamics obey
\begin{equation}
\dot{y} = -y + f(Wy,u_t,A) + g_{\mathrm{homeo}}(y),
\qquad
\dot{A} = -\lambda A + \Phi(y,u_t),
\label{eq:FHRL-cont}
\end{equation}
where the state $y$ represents the network’s internal activation, and $A$ is a fast-weight trace that records transient associations between recent inputs.
Unlike LNNs, FH-RL thus possesses two coupled time-scales within a single cell population:
a neuronal activity state $y$ and a fast-memory state $A$.

\medskip
The homeostatic term $g_{\mathrm{homeo}}(y)$ acts as a nonlinear radial controller that keeps the activity norm $\|y\|$ near unity.
Together with the leak term $-y$, it produces a Lyapunov-stable dynamics in which excitation and decay maintain a steady operating radius.
This mechanism serves a role similar to the state-dependent time constant in LNNs,
but it is derived explicitly from a homeostatic energy principle rather than learned through gradient optimization.

\medskip
Conceptually, LNNs emphasize temporal continuity and contraction of neural flows,
while FH-RL emphasizes reentrant refinement and stability under self-excitation.
The two frameworks share a continuous-time interpretation and a notion of state damping,
yet they differ in how memory and stability are implemented:

\begin{table}[h]
\centering
\renewcommand{\arraystretch}{1.3}
\begin{tabular}{p{3cm}p{5cm}p{5cm}}
\hline
\textbf{Feature} & \textbf{FH-RL Model} & \textbf{Liquid Neural Network} \\
\hline
Base equation &
$\dot{y} = -y + f(Wy,u_t,A) + g_{\mathrm{homeo}}(y), \quad
\dot{A} = -\lambda A + \Phi(y,u_t)$ &
$\dot{x} = -D(x,u)x + \sigma(Wx + Uu)$ \\

Stabilization mechanism &
Leak $(-y)$ + Homeostatic scaling $g_{\mathrm{homeo}}(y)$ &
State-dependent decay $D(x,u) > 0$ \\

Memory representation &
Fast-weight trace $A(t)$ (associative) &
Implicit in continuous state $x(t)$ \\

Learning focus &
Meta-stability of reentry and fast adaptation &
Adaptive time constants for temporal flows \\

Interpretation &
Dynamic attention and homeostatic self-regulation &
Adaptive continuous-time integration \\
\hline
\end{tabular}
\caption{Comparison between the FH-RL dynamics and Liquid Neural Networks.}
\label{tab:FHvsLNN}
\end{table}

\medskip
In summary, FH-RL extends the principles of LNNs by introducing an explicit homeostatic field and fast-weight memory,
thereby combining stability, reentrance, and short-term plasticity within one coherent continuous-time framework.
While LNNs approximate how biological neurons adapt their integration speeds,
FH-RL further models how cortical ensembles re-enter and self-stabilize during recurrent inference.
This makes the FH-RL dynamics a natural bridge between neural ODEs, liquid networks, and fast-weight transformer architectures.

\end{document}